\documentclass[letterpaper, 10 pt, conference]{ieeeconf}  % Comment this line out if you need a4paper

\usepackage{amsmath, amsfonts, amssymb} 
\usepackage{graphicx} 
\usepackage{color}       
\usepackage{array}       
\usepackage{bm}          
\usepackage{textcomp}   
\usepackage{stfloats}    
\usepackage{cite}       
\usepackage{url}         
\usepackage{lipsum}      
\usepackage{float}      
\usepackage{verbatim}   
\usepackage[utf8]{inputenc}  
\usepackage[letterpaper,top=2cm,bottom=2cm,left=2cm,right=2cm,marginparwidth=1.75cm]{geometry}  
\usepackage[T1]{fontenc} 
\usepackage[linesnumbered,ruled,vlined]{algorithm2e}   
\usepackage[caption=false,font=normalsize,labelfont=sf,textfont=sf]{subfig} 
\IEEEoverridecommandlockouts
\title{\LARGE \bf
Safety-Aware Robust Model Predictive Control for Robotic Arms in Dynamic Environments}

\author{Sanghyeon Nam$^{1}$, Dongmin Kim$^ {2}$, Seung-Hwan Choi$^{1}$, Chang-Hyun Kim$^{1}$, Hyoeun Kwon$^{1}$,\\ Hiroaki Kawamoto$^{3}$, and Suwoong Lee$^{1}$
\thanks{*This work was supported by the Industrial Technology Innovation Program(RS-2024-00424974, Robot and robot service digital twinization framework technology) funded by the Ministry of Trade, Industry Energy (MOTIE,Korea); in part by the Korea Institute of Industrial Technology as "Development of Core Technologies for a Working Partner Robot in the Manufacturing Field(kitech EO-25-0005)"; in part by the Korea Institute for Advancement of Technology (KIAT), funded by the Korea government (Ministry of Trade, Industry and Energy) (Project Number: RS-2024-00469384)
}% <-this % stops a space
\thanks{$^{1}$Korea Institute of Industrial Technology, Daegu 42994, Korea
        {\tt\small (nju1245, csw1496 , limition, hekwon525,  lee) @kitech.re.kr)}}%
\thanks{$^{2}$Daegu Mechatronics \& Materials Institute, Daegu 42714, Korea
        {\tt\small (dmkim@dmi.re.kr)}}%
\thanks{$^{3}$University of Tsukuba, Tsukuba 305-8577, Japan
        {\tt\small (kawamoto@iit.tsukuba.ac.jp)  }}%
}

\begin{document}

\maketitle
\thispagestyle{empty}
\pagestyle{empty}

%%%%%%%%%%%%%%%%%%%%%%%%%%%%%%%%%%%%%%%%%%%%%%%%%%%%%%%%%%%%%%%%%%%%%%%%%%%%%%%%
\section*{Abstract}
Robotic manipulators are essential for precise industrial pick-and-place operations, yet planning collision-free trajectories in dynamic environments remains challenging due to uncertainties such as sensor noise, and time-varying delays. Conventional control methods often fail under these conditions, motivating the development of Robust MPC (RMPC) strategies with constraint tightening. In this paper, we propose a novel RMPC framework that integrates phase-based nominal control with a robust safety mode, allowing smooth transitions between safe and nominal operations. Our approach dynamically adjusts constraints based on real-time predictions of moving obstacles—whether human, robot, or other dynamic objects—thus ensuring continuous, collision-free operation. Simulation studies demonstrate that our controller improves both motion naturalness and safety, achieving faster task completion than conventional methods.

\section{Introduction}

Robotic manipulators have become an essential part of modern industrial and service applications, enhancing productivity and reducing human risk in complex tasks. Their ability to perform precise pick-and-place operations in hazardous environments has driven extensive research into advanced control strategies \cite{al2023generalized}. In particular, collision-free trajectory planning is a fundamental problem that directly impacts the safety and efficiency of robotic arms, especially when navigating through environments with both static and dynamic obstacles \cite{surati2021pick}.

Recent trends in robotic arm trajectory planning have explored various approaches, including graph-based and tree-based spatial planning methods \cite{dai2022review} as well as energy optimization techniques that compare classical PID with model predictive control (MPC)-based controllers \cite{zahaf2022robust}. Advanced MPC schemes, which integrate feedback linearization to manage system nonlinearities \cite{mishra2024model}, have shown particular promise. Notably, the dynamic MPC controller developed in \cite{killpack2016model} addresses the shortcomings of quasi-static MPC methods that fail to capture dynamic phenomena such as joint inertia and collision impact forces during rapid movements. By employing multi-time-step predictions and modeling dynamic effects, this approach enables faster reaching speeds and more robust control in cluttered environments, while effectively regulating contact forces. Experimental results demonstrate that the dynamic MPC framework significantly enhances both performance and safety in robot arm control, making it a strong candidate for complex manipulation tasks.

Despite the progress made by conventional MPC methods in various applications, they still struggle to capture critical uncertainties—such as sensor noise, and time-varying delays—in human-robot collaborative environments. Recent efforts have sought to enhance MPC by integrating neural network-based tuning and heuristic optimization techniques \cite{elsisi2021effective, elsisi2021improved, guruji2016time, fu2018improved}. However, these learning-based approaches are often prone to converging to local optima and require extensive parameter tuning and training data, which limits their ability to rapidly adapt to unforeseen environmental changes and maintain robust performance over longer prediction horizons. In contrast, Robust MPC (RMPC) explicitly addresses uncertainties by incorporating strategies such as constraint tightening and robust optimization techniques \cite{zahaf2022robust}. Moreover, recent work has improved constraint handling in MPC formulations \cite{sleiman2021constraint}, demonstrating that RMPC-based approaches offer a more reliable and effective solution for controlling robotic arms under uncertainty.

To further address the high computational burden associated with continuous long-horizon predictive control, it is essential to divide the control strategy into distinct modes: a nominal phase-based trajectory for regular operation and an RMPC safety mode for managing sudden obstacles or human interactions \cite{gold2022model}. This mode-switching approach ensures that short-term, real-time predictions accurately capture the system's dynamics while maintaining overall efficiency and safety.

In this paper, we propose a novel RMPC framework that integrates conventional phase-based control with a robust safety mode. Our approach explicitly considers uncertainties such as sensor measurement errors, and it implements a smooth transition mechanism between nominal and RMPC modes. By leveraging recent advances in safe collaborative robotic arm control \cite{khoramshahi2023practical}, our method offers a robust and energy-efficient solution for robotic manipulators operating in dynamic and uncertain environments.

The main contributions of this work are as follows:
\begin{enumerate}
    \item We develop an integrated framework that merges phase-based nominal control with an RMPC safety mode to ensure collision-free trajectory tracking in dynamic environments while reducing computational load.
    \item We employ constraint tightening techniques that incorporate sensor uncertainties, ensuring robust performance even under worst-case scenarios.
    \item We validate our approach with simulation studies that demonstrate improved motion naturalness and safety during pick-and-place tasks.
\end{enumerate}

\section{Robotic Arm Modeling}

\begin{figure}[t]
    \centering
    \includegraphics[width=55mm]{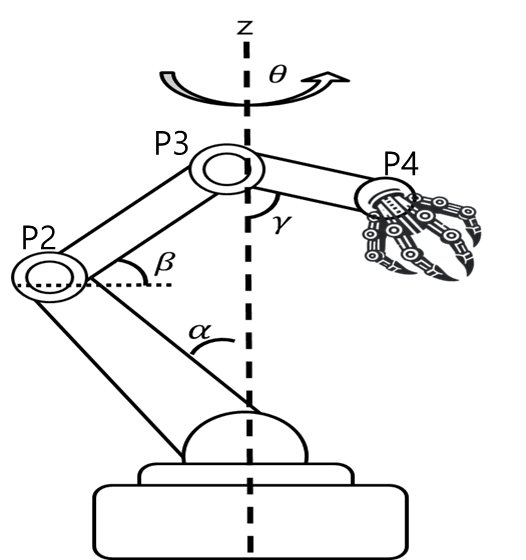}
    \caption{Robot arm structure and joint configuration.}
    \label{Archi}
\end{figure}

The robotic arm used in this study is a 4-DOF manipulator consisting of three joint angles, \(\alpha, \beta, \gamma\), which define the configuration of the arm, and a rotational parameter \(\theta\) that represents the base rotation about the z-axis, as depicted in Fig.~\ref{Archi}. The arm operates within a predefined range and is subject to kinematic and dynamic constraints.

\subsection{Kinematics}

The position of key points along the robotic arm is determined by the joint angles and link lengths. Given the link lengths \( L_1, L_2, L_3 \), the positions of significant points are derived as follows:

\textbf{End of \(L_1\) (P2):}
\begin{align}
    x_2 &= -L_1 \sin\alpha \cos\theta, \nonumber \\
    y_2 &= -L_1 \sin\alpha \sin\theta,  \nonumber \\
    z_2 &= L_1 \cos\alpha.
\end{align}

\textbf{End of \(L_2\) (P3):}
\begin{align}
    x_3 &= x_2 + L_2 \cos\beta \cos\theta,  \nonumber \\
    y_3 &= y_2 + L_2 \cos\beta \sin\theta, \nonumber  \\
    z_3 &= z_2 + L_2 \sin\beta.
\end{align}

\textbf{End of \(L_3\) (P4, End-Effector):}
\begin{align}
    x_4 &= x_3 + L_3 \sin\gamma \cos\theta, \nonumber  \\
    y_4 &= y_3 + L_3 \sin\gamma \sin\theta, \nonumber  \\
    z_4 &= z_3 - L_3 \cos\gamma.
\end{align}

\subsection{System Dynamics}

The robotic arm follows nonlinear system dynamics represented by:
\begin{equation}
    \mathbf{x}_{k+1} = f(\mathbf{x}_k, \mathbf{u}_k),
\end{equation}
where the control input vector is
\begin{equation}
    \mathbf{u}_k = \begin{bmatrix} \dot{\alpha}_k & \dot{\beta}_k & \dot{\gamma}_k & \dot{\theta}_k \end{bmatrix}^\top.
\end{equation}
The angular velocities in \(\mathbf{u}_k\) define the rates of change for each joint and the base rotation.

Using Euler integration, the state update is given by:
\begin{align}
    \alpha_{k+1} &= \alpha_k + \dot{\alpha}_k\,\Delta t, \\
    \beta_{k+1}  &= \beta_k + \dot{\beta}_k\,\Delta t, \\
    \gamma_{k+1} &= \gamma_k + \dot{\gamma}_k\,\Delta t, \\
    \theta_{k+1} &= \theta_k + \dot{\theta}_k\,\Delta t.
\end{align}

\subsection{Constraints}

The robotic arm operates under physical constraints to ensure safe motion:

\begin{itemize}
    \item \textbf{Joint Angle Constraints}:
    \begin{align}
        & \alpha_{\min} \leq \alpha_k \leq \alpha_{\max}, \label{(11)} \\
        & \beta_{\min} \leq \beta_k \leq \beta_{\max}, \\
        & \gamma_{\min} \leq \gamma_k \leq \gamma_{\max}.
    \end{align}
    
    \item \textbf{Base Rotation Constraints}:
    \begin{equation}
        \theta_{\min} \leq \theta_k \leq \theta_{\max}.
        \label{(14)}
    \end{equation}
    
    \item \textbf{Control Input Constraints}:
    \begin{align}
        & \dot{\alpha}_{\min} \leq \dot{\alpha}_k \leq \dot{\alpha}_{\max}, \label{Cinput14} \\
        & \dot{\beta}_{\min} \leq \dot{\beta}_k \leq \dot{\beta}_{\max}, \\
        & \dot{\gamma}_{\min} \leq \dot{\gamma}_k \leq \dot{\gamma}_{\max}, \\
        & \dot{\theta}_{\min} \leq \dot{\theta}_k \leq \dot{\theta}_{\max}. \label{Cinput17}
    \end{align}
\end{itemize}

\section{Control Design}
\begin{figure}[t]
    \centering
    \includegraphics[width=65mm]{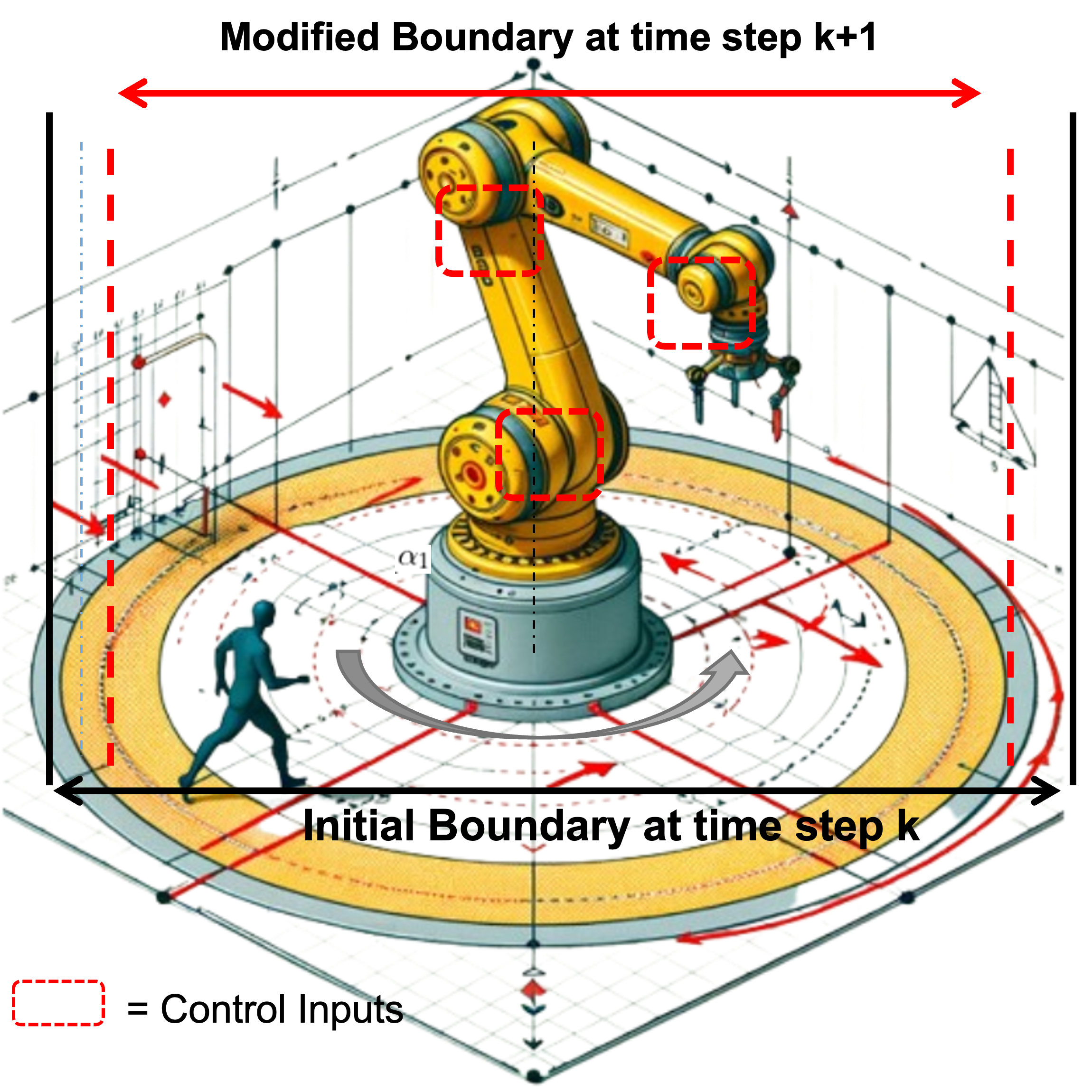}
    \caption{Scenario Illustration: A Man Approaching an Operating Robotic Arm.}
    \label{ExampleFigure}
\end{figure}

In this section, we present the optimization process for robot arm control using Robust MPC, detailing the control design and the method for determining a robust invariant set. This ensures safe operation without unnecessary stops or reductions in velocity during iterative work cycles. The considered scenario is illustrated in Fig.~\ref{ExampleFigure}, where workers or other robots operating near the robotic arm are detected by its sensors, requiring the arm to actively avoid them. In this scenario, the robotic arm continuously performs repetitive pick-and-place tasks in an environment without physical fences, where various humans and robots move unpredictably. The objective is to ensure smooth task execution by effectively avoiding these dynamic obstacles while maintaining operational efficiency.

\subsection{Feasible Set and Constraint Definitions}

The overall feasible set of state and input pairs is defined as:
\begin{equation}
    \mathcal{R} = \{ (\bm{x}, \bm{u}) \in \mathbb{R}^n \times \mathbb{R}^m \mid \bm{x} \in \mathcal{X},\, \bm{u} \in \mathcal{U} \},
\end{equation}
where \(\mathcal{X}\) is the feasible state set and \(\mathcal{U}\) is the feasible input set. In particular, the feasible state set is given by:
\begin{equation}
    \mathcal{X} = \left\{ \bm{x} \in \mathbb{R}^n \, \middle| \,
    \begin{aligned}
        &x^2 + y^2 \leq R_{arm}^2, \\
        &z_{\text{min}} \leq z \leq z_{\text{max}}, \\
        &\text{Constraints given in} (\ref{(11)}) \text{--} (\ref{(14)})
    \end{aligned}
    \right\}.
    \label{Constraint_origin}
\end{equation}
In this context, the value \(R_{\text{arm}}\) denotes the constraint for turning radius of the robotic arm, including its safety margin, while \(x\), \(y\), and \(z\) denote the \(x\)-coordinate, \(y\)-coordinate, and vertical position of point 4, respectively.

\subsection{Disturbance Set Definition}
An approaching object's predicted path is treated as a disturbance, effectively reducing the set of feasible states. To model the uncertainties in object motion (both velocity and height), we define the disturbance set as a product set:
\begin{equation}
    \mathcal{W} = \left\{ \bm{w} \in \mathbb{R}^2 \,\middle|\, |w_v| \leq \delta_v,\quad |w_z| \leq \delta_z \right\},
\end{equation}
where \(w_v\) and \(w_z\) denote the uncertainties in velocity and height, respectively, with bounds \(\delta_v\) and \(\delta_z\).

To ensure compatibility with the state constraint set $\mathcal{X}$, we redefine the disturbance set in a three-dimensional space by transforming the velocity uncertainty \(w_v\) into a radial displacement \(w_r\):
\begin{equation}
    w_r = w_v \cdot \Delta t.
\end{equation}
Thus, the new disturbance set \( \mathcal{W}' \) in \(\mathbb{R}^3\) is given by:
\begin{equation}
    \mathcal{W}' = \left\{ \bm{w} \in \mathbb{R}^3 \,\middle|\, |w_r| \leq \delta_r,\quad |w_v| \leq \delta_v, \quad |w_z| \leq \delta_z \right\},
    \label{disturbance}
\end{equation}
where \(\delta_r\) represents the maximum radial displacement due to velocity uncertainty.

\subsection{Robust Invariant Set Based Phase Trajectory}

The robust invariant set \(\mathcal{Z}\) is defined as the set of states in the nominal constraint set \(\mathcal{X}\) that remain feasible for all bounded disturbances. This ensures that phase mode trajectories are confined within \(\mathcal{Z}\), so that if the system switches to RMPC mode, even worst-case disturbances will not drive the state out of the safe region:
\begin{equation}
\begin{split}
    \mathcal{Z} = \Big\{ \bm{x} \in \mathcal{X} \;\Big|\; &\forall\, \bm{w} \in \mathcal{W},\, \exists\, \bm{u} \in \mathcal{U}\text{ s.t.} \\
    &f(\bm{x},\bm{u}) + \bm{w} \in \mathcal{X} \Big\}.
\end{split}
\end{equation}

This set is computed via backward reachability analysis from a worst-case state (e.g., the robot arm at maximum extension corresponding to maximum object height interference). For a finite prediction horizon \(N_p\), the backward reachable set at time \(k\) is given by:
\begin{equation}
\begin{split}
    \mathcal{Z}_k = \Big\{ \bm{x} \in \mathcal{X} \,\Big|\; &\exists \{\bm{u}_i\}_{i=k}^{k+N_{p}-1} \subset \mathcal{U} \text{ s.t.} \\
    &\bm{x}_{i+1} = f(\bm{x}_i,\bm{u}_i) + \bm{w}_i,\quad \forall\, \bm{w}_i \in \mathcal{W}, \\
    &\bm{x}_{k+N_p} \in \mathcal{X} \Big\},
\end{split}
\end{equation}
for \(k = 0,1,\dots,N_{p}-1\). This formulation guarantees that, despite uncertainties, the system can be steered into the safe set \(\mathcal{Z}\) within \(N_p\) steps.

\subsection{Constraint Tightening for object Motion}

To ensure robust feasibility under uncertainties introduced by object motion, the RMPC framework dynamically tightens constraints. The feasible state set is adjusted based on the Pontryagin difference:
\begin{equation}
    \mathcal{X}_r = \mathcal{X} \ominus \mathcal{W'}. \label{setdiff}
\end{equation}

When RMPC is activated, the constraints in (\ref{Constraint_origin}) are ultimately tightened as follows:
\begin{align}
    x^2 + y^2 &\leq \Bigl(R_{arm} - w_r\Bigr)^2, \label{eq:mod_xy}\\
    z &\geq z_{\text{min}} + w_z - \epsilon. \label{eq:mod_zmin}
\end{align}
Here, the value $\epsilon$ is introduced to allow minor constraint violations during mode transitions; its detailed role is discussed in the next subsection.

\begin{algorithm}[t]
\caption{Adaptive RMPC Constraint Adjustment via Pontryagin Difference}
\label{algorithm}
\SetAlgoLined
\KwIn{
  $p_h(t)$: Measured object position\\
  $v_h(t)$: Measured object velocity\\
  $p_4(t)$: Robot end-effector position
}
\KwOut{ $\mathcal{X}_{r}$: Adjusted state constraint set }
\textbf{Initialization:}\\
Set control set $\mathcal{Z}$, nominal constraints $\mathcal{X}$, prediction horizon $N_p$, and uncertainty mapping function $h(\cdot)$;\;
\While{robot is operating}{
    Measure $p_h(t)$, $v_h(t)$, and $p_4(t)$;\;
    Predict future positions $\hat{p}_h(t+i\Delta t)$ for $i=1,\ldots,N_p$;\;
    \For{$i=1$ to $N_p$}{
        Compute $d_i = \|\hat{p}_h(t+i\Delta t) - p_4(t)\|$;\;
    }
    Let $d_{\min} = \min\{d_1,\ldots,d_{N_p}\}$;\;
    \eIf{an imminent collision is predicted}{
        Activate RMPC mode;\;
        Compute the uncertainty set $\mathcal{W}' = h(p_h(t), v_h(t), w_v(t), w_r(t), w_z(t))$;\;
        Set $\mathcal{X}_{r} = \mathcal{X} \ominus \mathcal{W}'$;\;
        Solve RMPC with $\mathcal{X}_{r}$ and apply control;\;
    }
    {
        Continue nominal control;\;
    }
}
\Return{$\mathcal{X}_{r}$}\;
\end{algorithm}

\subsection{Optimization Formulation}

The RMPC problem is formulated as a nonlinear optimization problem that minimizes a cost function while satisfying dynamically tightened constraints and accounting for bounded disturbances. The cost function is defined as
\begin{equation}
    J = \sum_{k=0}^{N_{p}-1} \Big( c_1\, |u_{k,1}| + c_2\, |u_{k,2}| + c_3\, |u_{k,3}| + c_4\, \epsilon_{k} \Big),
\end{equation}
where \(u_{k,i}\) denotes the \(i\)th component of the control input at time step \(k\), and \(c_1, c_2, c_3, c_4\) are weighting coefficients (e.g., \(c_1=1\), \(c_2=3\), \(c_3=5\), \(c_4=100\)).

Finally, we formulate the following optimization problem that incorporates tightened constraints to ensure safety:
\begin{align}
    \min_{\bm{u}_k \in \mathcal{U} }\quad & J = \sum_{k=0}^{N_{P}-1} \Big( c_1\, |u_{k,1}| + c_2\, |u_{k,2}| + c_3\, |u_{k,3}| + c_4\, \epsilon_k \Big), \nonumber \\
    \text{s.t.} \quad & \bm{x}_{k+1} = f(\bm{x}_k,\bm{u}_k), \nonumber \\
    & \bm{x}_k \in \mathcal{X}_r(k),  \nonumber \\
    & (\ref{Cinput14})-(\ref{Cinput17}), \nonumber  \\
    & 0 \leq \epsilon_k \leq 3, \nonumber \\ & \forall k = 0, \dots , N_{p}-1. 
\end{align}
The variable \(\epsilon_k\) is introduced as a slack variable for the state constraints. A high penalty coefficient \(c_4\) is applied to \(\epsilon_k\) in the cost function to minimize any deviation from the tightened constraints, thereby enabling a soft constraint formulation that allows the system to quickly satisfy the constraints once the RMPC mode is activated, even under worst-case disturbances. Furthermore, the dynamic set \(\mathcal{X}_r(k)\) captures the time-varying nature of the constraints—for example, ensuring that the end-effector's \(z\)-coordinate, adjusted by the slack variable, remains above a specified height at each time step. This guarantees that, despite the nonlinear model and external disturbances, the RMPC drives the system into the safe operating region.
The overall process of the proposed control framework for the robotic arm is outlined in Algorithm \ref{algorithm}.

\begin{figure*}[t]
    \centering
    \begin{minipage}{0.3\textwidth}
        \centering
        \fbox{\includegraphics[width=\textwidth]{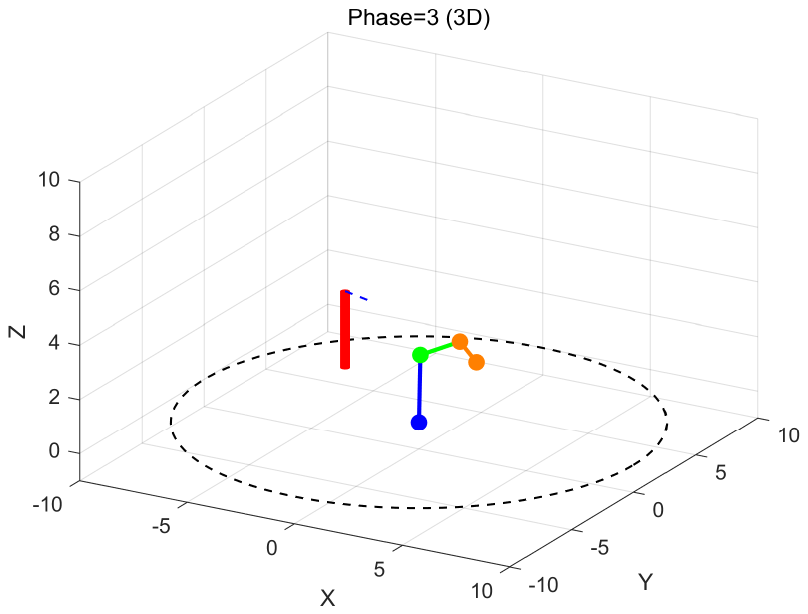}}
    \end{minipage}
    \begin{minipage}{0.3\textwidth}
        \centering
        \fbox{\includegraphics[width=\textwidth]{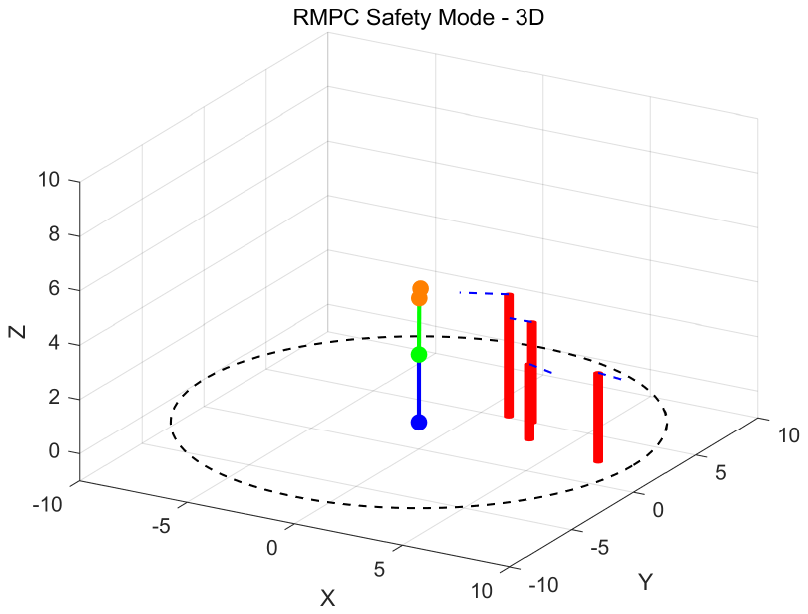}}
    \end{minipage}
    \begin{minipage}{0.3\textwidth}
        \centering
        \fbox{\includegraphics[width=\textwidth]{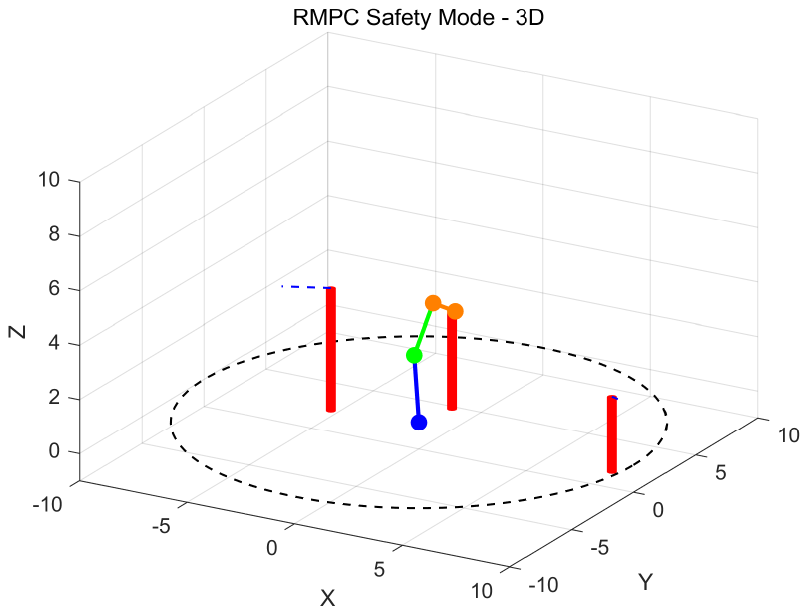}}
    \end{minipage}
        \begin{minipage}{0.3\textwidth}
        \centering
        \fbox{\includegraphics[width=\textwidth]{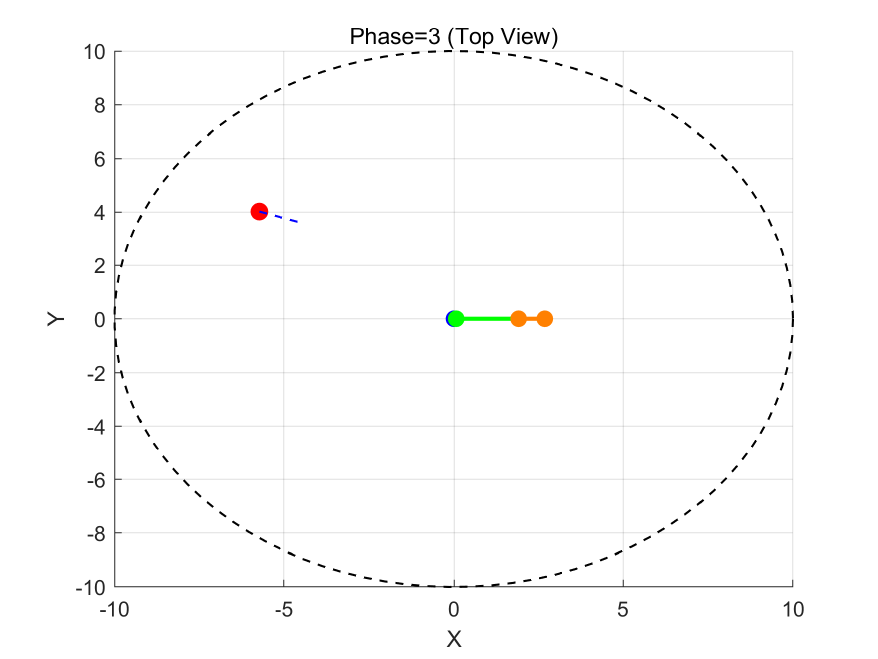}}\\
        {\small (1)}
    \end{minipage}
    \begin{minipage}{0.3\textwidth}
        \centering
        \fbox{\includegraphics[width=\textwidth]{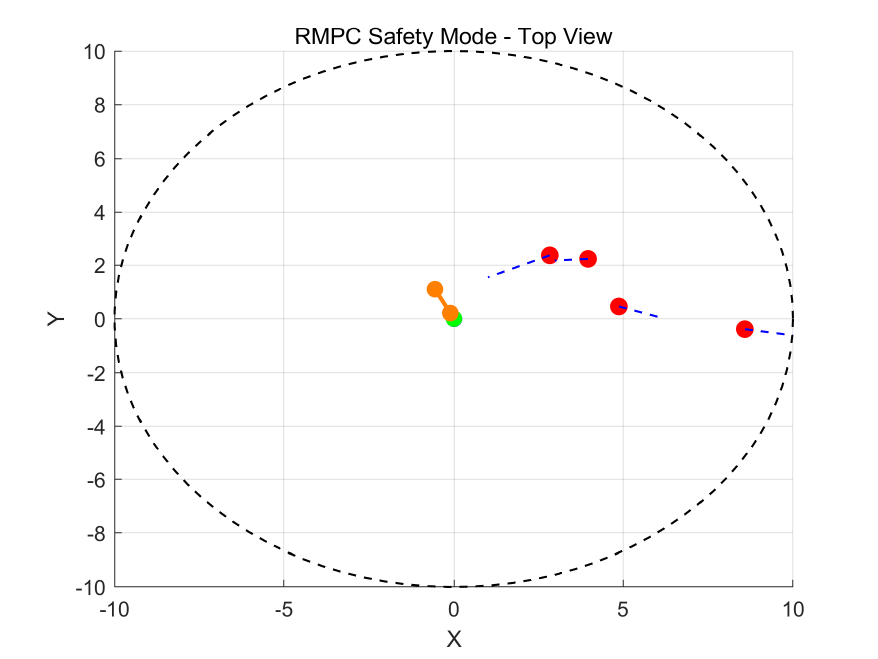}}\\
        {\small (2)}
    \end{minipage}
    \begin{minipage}{0.3\textwidth}
        \centering
        \fbox{\includegraphics[width=\textwidth]{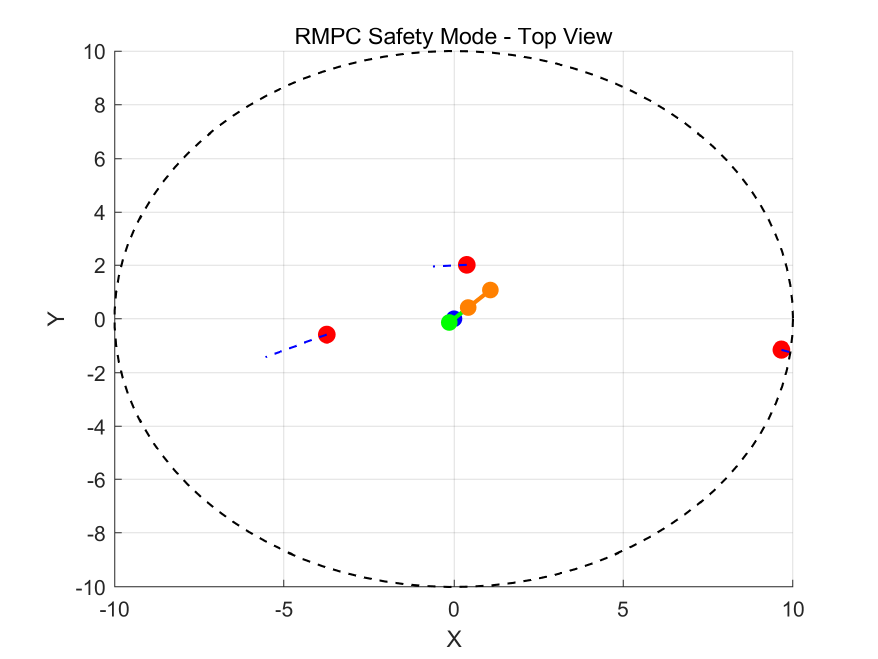}}\\
        {\small (3)}
    \end{minipage}
    \caption{Simulation captures of frames.}
    \label{3D}
\end{figure*}

\section{Simulation}
In this section, we conduct MATLAB simulation studies that incorporate object-generation modeling to replicate dynamic environments.

\subsection{Object Generation}
To account for dynamic environments, approaching object motion is modeled as linear trajectories based on velocity and direction. The position of an object at time \(t\) is given by:
\begin{equation}
    \bm{p}(t) = \bm{p}_0 + \bm{v} t,
\end{equation}
where:
\begin{itemize}
    \item \( \bm{p}_0 = [x_0, y_0]^T \) is the initial position within the detectable circle.
    \item \( \bm{v} = v \cdot [\cos(\theta), \sin(\theta)]^T \) is the velocity vector with magnitude \( v \) and direction \( \theta \).
\end{itemize}

The robotic arm continuously receives real-time measurements of object height, position, and speed. In each sample time, the measurement data include uncertainties in velocity and height, denoted by \(w_v\) and \(w_z\), respectively, with corresponding bounds \(\delta_v\) and \(\delta_z\) addressed in (20).

\subsection{Case Study}

The simulations were conducted on an Intel Core i7-12700K processor (12 cores, 20 threads, base clock 3.6GHz, max turbo 5.0GHz) in a MATLAB environment, where a nonlinear solver was utilized to compute the RMPC. Throughout the RMPC activation period, the computation time per iteration ranged from 0.008s to 0.052s, with an average of 0.032s, effectively demonstrating the feasibility of real-time implementation.

Fig.~\ref{3D} presents snapshots from our simulation, where the upper images depict the side view and the lower images show the top view, both captured at the same time instants. In these figures, the robotic arm is located at the center of the detectable region (dotted circle), while red cylinders indicate approaching objects. Each object is generated with random heights and velocities, demonstrating the system’s ability to handle varying levels of uncertainty.

\begin{figure*}[t]
    \centering
    \includegraphics[width=184mm]{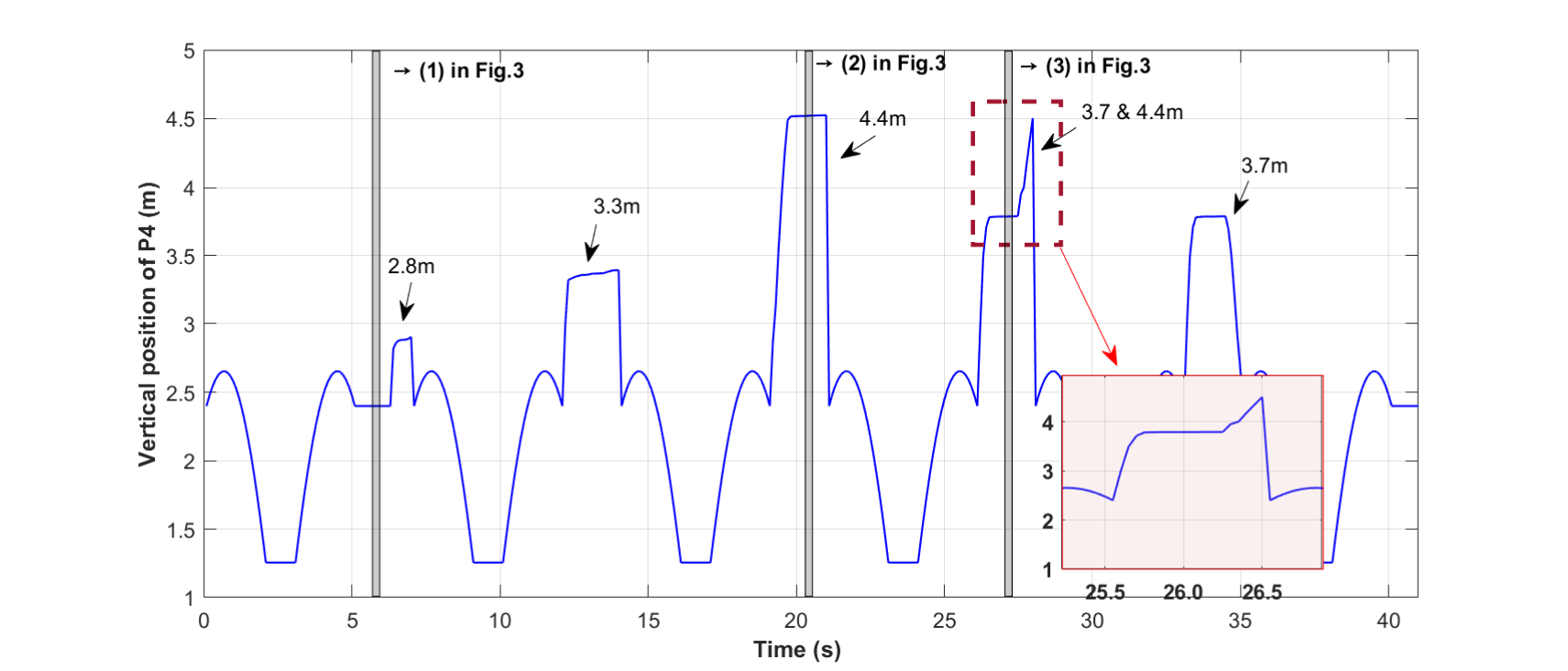}
    \caption{Vertical Position Changes of Point 4 Over Time.}
    \label{float}
\end{figure*}

Fig.~\ref{float} illustrates the full trajectory of the end-effector’s vertical position (\(P_4\)) throughout the simulation. In this figure, the labels (1), (2), and (3) correspond to the captured snapshots at those respective time points as shown in Fig.~\ref{3D}. Despite the sensing uncertainties that arise at each sampling instant, no undesirable oscillations occur when the arm adjusts its height. This observation confirms that the uncertainty-aware constraint tightening effectively mitigates abrupt movements, enabling a stable yet adaptive response for safe robotic manipulation.

In scenarios with higher object densities—particularly frames (2) and (3) in Fig.~\ref{3D}—the controller accounts for the maximum object height (e.g., 2.8\,m, 3.3\,m, 3.7\,m, and 4.4\,m). As depicted in Fig.~\ref{float}, the vertical position of the end-effector is adjusted based on future predictions, allowing the controller to adapt from targeting a 3.7\,m object to a 4.4\,m object while in motion. The numerical changes in the end-effector’s height are highlighted in the enlarged red dashed box in Fig.~\ref{float}.

When the environment is deemed safe, the controller transitions smoothly from RMPC mode back to its nominal trajectory-following mode to reduce computational overhead. This switch is evident in frame (15) of Fig.~\ref{3D}, where the arm descends from an elevated position once the potential collision threat is no longer present. Overall, these results validate that the proposed approach not only ensures collision avoidance but also maintains stable motion under dynamic and uncertain conditions.

To demonstrate the efficiency of our proposed controller, we imported the control method from \cite{zeng2013mobile} as a baseline. Although this method is not state-of-the-art, it is widely applied in real-world scenarios to ensure worker safety by reducing speed and halting operation when a object approaches. In our simulation, the primary metric is the Euclidean distance 
\begin{equation}
    d = \|\bm{p}_r - \bm{p}_h\|, \quad \bm{p}_r, \bm{p}_h \in \mathbb{R}^2,
    \label{eq:distance}
\end{equation}
which is used to define three safety regions:
\begin{itemize}
    \item \textbf{Safe Region (\(d > R_2\))}: The robot moves at its nominal velocity.
    \item \textbf{Active Region (\(R_3 \leq d \leq R_2\))}: The robot’s velocity is reduced, and a repulsive force modifies its trajectory:
    \begin{equation}
        \bm{F}_{\text{repulsive}} = -k \frac{\bm{p}_r - \bm{p}_h}{\|\bm{p}_r - \bm{p}_h\|^3},
        \label{eq:repulsive_force}
    \end{equation}
    \item \textbf{Critical Region (\(d < R_3\))}: The robot stops to avoid collision.
\end{itemize}

Accordingly, the robot’s velocity is updated as:
\begin{equation}
    v_r =
    \begin{cases}
        v_{\text{nominal}}, & \text{if } d > R_2, \\
        v_{\text{adjusted}}, & \text{if } R_3 \leq d \leq R_2, \\
        0, & \text{if } d < R_3.
    \end{cases}
    \label{eq:velocity_update}
\end{equation}

We simulated this controller under the same scenario depicted in Fig.~\ref{3D}, where all objects are generated simultaneously at fixed positions and follow identical trajectories throughout the simulation. Our proposed controller completed the task in 41.46 seconds, whereas the imported controller required 68.75 seconds. Notably, our method maintained continuous motion without unnecessary stops, thanks to its proactive constraint tightening that prevents speed reductions even in crowded settings. In contrast, the imported controller frequently halted when objects came too close, leading to inefficient task execution. These results underscore that our approach not only ensures safety but also significantly improves operational efficiency. We expect that our controller's advantages will become even more pronounced in more crowded scenarios, and in future work we plan to test it under a wider range of conditions and specific cases to further validate its versatility. Due to page limitations, a detailed quantitative comparison with state-of-the-art methods is not included here; however, extensive performance evaluations will be presented in future studies.

\section{Conclusion}
This study proposed an RMPC-based control framework for robotic arms in dynamic object-interactive environments, ensuring both safety and smooth motion. By integrating phase-based nominal control with a robust safety mode, the approach effectively handles uncertainties like sensor noise and delays achieving collision-free trajectory tracking. Simulation results demonstrate improved motion naturalness and safety over conventional MPC-based controllers.

Due to page constraints, this study primarily focused on presenting the core control framework rather than conducting an extensive comparison with state-of-the-art algorithms. Additionally, a detailed analysis of various quantitative performance metrics was not included. In future work, we plan to extend this research into a journal publication by systematically comparing our approach against the latest control methods to quantitatively assess its advantages in terms of safety and operational efficiency.  

Furthermore, we aim to expand the scope of our simulation scenarios to validate the controller's adaptability under a wider range of dynamic conditions, demonstrating its ability to maintain efficient motion across diverse environments. Another key area of improvement will involve refining the cost function beyond its current simplistic form. By incorporating scenario-specific energy optimization terms, we intend to enhance overall efficiency while ensuring robust constraint satisfaction. These advances will contribute to the development of a more versatile and practically deployable RMPC framework for robotic applications in human-interactive settings.


%%%%%%%%%%%%%%%%%%%%%%%%%%%%%%%%%%%%%%%%%%%%%%%%%%%%%%%%%%%%%%%%%%%%%%%%%%%%%%%%

%%%%%%%%%%%%%%%%%%%%%%%%%%%%%%%%%%%%%%%%%%%%%%%%%%%%%%%%%%%%%%%%%%%%%%%%%%%%%%%%

\bibliography{reference}    % reference
\bibliographystyle{IEEEtran}

\end{document}